\documentclass[journal]{IEEEtran}
\usepackage{amsfonts}
\usepackage{lineno,hyperref}
\usepackage{times}
\usepackage{epsfig}
\usepackage{graphicx}
\usepackage{amsmath}
\usepackage{amssymb}
\usepackage{amstext}
\usepackage{multirow}
\usepackage{booktabs}
\usepackage{color}
\modulolinenumbers[5]

\usepackage{ifpdf}

\usepackage{cite}
\usepackage{amsmath}
\usepackage{algorithmic}

\usepackage{array}
\usepackage{url}

\begin{document}
%
\title{DeepSkeleton: Learning Multi-task Scale-associated Deep Side Outputs for Object Skeleton Extraction in Natural Images}
%
%
%

\author{Wei Shen,
        Kai Zhao,
        Yuan Jiang, Yan Wang, Xiang Bai and Alan Yuille
\thanks{W. Shen, K. Zhao and Y. Jiang are with Key Laboratory of Specialty Fiber Optics and Optical Access Networks, Shanghai University, Shanghai 200444
China. W. Shen is also with Department of Computer Science, Johns Hopkins University, Baltimore, MD 21218-2608 USA. E-mail: shenwei1231@gmail.com, \{zeakey,jy9387\}@outlook.com.}
\thanks{Y. Wang is with Department of Computer Science, Johns Hopkins University, Baltimore, MD 21218-2608 USA. E-mail: wyanny.9@gmail.com}
\thanks{X. Bai is with School of Electronic Information and Communications, Huazhong University of Science and Technology, Wuhan 430074 China. Email: xiang.bai@gmail.com}
\thanks{A. Yuille is with Department of Computer Science, Johns Hopkins University, Baltimore, MD 21218-2608 USA. E-mail: alan.yuille@jhu.edu .}}

%
%

\markboth{}%
{Shell \MakeLowercase{\textit{et al.}}: Bare Demo of IEEEtran.cls for IEEE Journals}
%



\maketitle

\begin{abstract}
Object skeletons are useful for object representation and object detection. They are complementary to the object contour, and provide extra information, such as how object scale (thickness) varies among object parts. But object skeleton extraction from natural images is very challenging, because it requires the extractor to be able to capture both local and non-local image context in order to determine the scale of each skeleton pixel. In this paper, we present a novel fully convolutional network with multiple scale-associated side outputs to address this problem. By observing the relationship between the receptive field sizes of the different layers in the network and the skeleton scales they can capture, we introduce two scale-associated side outputs to each stage of the network. The network is trained by multi-task learning, where one task is skeleton localization to classify whether a pixel is a skeleton pixel or not, and the other is skeleton scale prediction to regress the scale of each skeleton pixel. Supervision is imposed at different stages by guiding the scale-associated side outputs toward the groundtruth skeletons at the appropriate scales. The responses of the multiple scale-associated side outputs are then fused in a scale-specific way to detect skeleton pixels using multiple scales effectively. Our method achieves promising results on two skeleton extraction datasets, and significantly outperforms other competitors. Additionally, the usefulness of the obtained skeletons and scales (thickness) are verified on two object detection applications: Foreground object segmentation and object proposal detection.
\end{abstract}


%
\IEEEpeerreviewmaketitle

\section{Introduction} \label{sec:intro}
\begin{figure}[!h]
\centering
\includegraphics[width=1.0\linewidth]{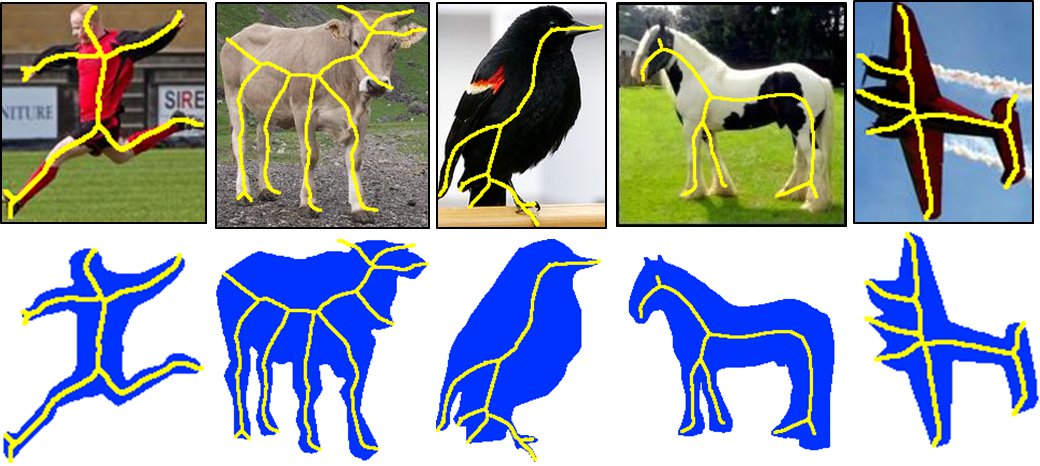}
\caption{Object skeleton extraction in natural images. The skeletons are in yellow. Top: Skeleton localization. Bottom: Scale prediction which enables object segmentation (blue regions are the segments reconstructed from skeletons according to the scales).}\label{fig:sk_ex}
\end{figure}

In this paper, we investigate an important and nontrivial problem in computer vision, namely object skeleton extraction from natural images (Fig.~\ref{fig:sk_ex}). Here, the concept of ``object'' means a standalone entity with a well-defined boundary and center~\cite{Ref:AlexeDF10}, such as an animal, a human, and a plane, as opposed to amorphous background stuff, such as sky, grass, and mountain. The skeleton, also called the \emph{symmetry axis}, is a useful structure-based object descriptor. Extracting object skeletons directly from natural images can deliver important information about the presence and size of objects. Therefore, it is useful for many real applications including object recognition/detection~\cite{Ref:Bai09,Ref:TrinhK11}, text recognition~\cite{Ref:Zhang15}, road detection and blood vessel detection~\cite{Ref:SironiLF14}.

Skeleton extraction from pre-segmented images~\cite{Ref:Saha15} has been well studied and successfully applied to shape-based object matching and recognition~\cite{Ref:Siddiqi99,Ref:Sebastian04,Ref:Demirci06,Ref:Bai08}. However, such methods have severe limitations when applied to natural images, because segmentation from natural images is still an unsolved problem.

Skeleton extraction from natural images is a very challenging problem, which requires addressing two tasks. One is skeleton localization to classify whether a pixel is a skeleton pixel or not (the top row in Fig.~\ref{fig:sk_ex}) and the other is skeleton scale prediction to estimate the scale of each skeleton pixel (the bottom row in Fig.~\ref{fig:sk_ex}). The latter task has not been studied explicitly in the past, although it is very important, because using the predicted scales, we can obtain object segmentation from a skeleton directly. In this paper, we address skeleton localization and scale prediction in a unified framework which performs them simultaneously. The main difficulties for skeleton extraction stem from four issues: (1) The complexity of natural scenes: Natural scenes are typically very cluttered. Amorphous background elements, such as fences, bricks and even the shadows of objects, exhibit some self-symmetry, and thus can cause distractions. (2) The diversity of object appearance: Objects in natural images exhibit very different colors, textures, shapes and sizes. (3) The variability of skeletons: local skeleton segments have a variety of patterns, such as straight lines, T-junctions and Y-junctions. (4) The \emph{unknown-scale problem}: A local skeleton segment is naturally associated with an unknown scale, determined by the thickness of its corresponding object part. We term this last problem the unknown-scale problem for skeleton extraction.

A number of methods have been proposed to perform skeleton extraction or skeleton localization in the past decade. Broadly speaking, they can be categorized into two groups: (1) Traditional image processing methods~\cite{Ref:Yu04,Ref:Jang01,Ref:Lindeberg98,Ref:ZhangC07}, which compute skeletons from a gradient intensity map according to some geometric constraints between edges and skeletons. Due to the lack of
supervised learning, these methods have difficulty in handling images with complex scenes; (2) Recent learning based methods~\cite{Ref:Tsogkas12,Ref:Levinshtein09,Ref:Lee13,Ref:SironiLF14,Ref:Widynski14}, which learn a per-pixel classification or segment-linking model based on hand-designed features for skeleton extraction computed at multi-scales. But the limitations of hand-designed features cause these methods to fail to extract the skeletons of objects with complex structures and cluttered interior textures. In addition, such per-pixel/segment models are usually time consuming. More importantly, most current methods only focus on skeleton localization, but are unable to predict skeleton scales, or are only able to provide a coarse prediction for skeleton scales. This big shortcoming limits the application of the extracted skeletons to object detection. Consequently, there remain big gaps between these skeleton extraction methods and human perception, in both performance and speed. Skeleton extraction has the unique aspect of requiring both local and non-local image context, which requires new techniques for both multi-scale feature learning and classifier learning. This is challenging, since visual complexity increases exponentially with the size of the context field.

To tackle the obstacles mentioned above, we develop a holistically-nested network with multiple scale-associated side outputs for skeleton extraction. The holistically-nested network (HED)~\cite{Ref:Xie15} is a deep fully convolutional network (FCN)~\cite{Ref:LongSD15}, which enables holistic image training and prediction for per-pixel tasks. A side output is the output of a hidden layer of a deep network. The side outputs of the hidden layers, from shallow to deep, give multi-scale responses, and can be guided by supervision to improve the directness and transparency of the hidden layer learning process~\cite{Ref:Lee15}. Here we connect two sibling scale-associated side outputs to each convolutional layer in the holistically-nested network to address the unknown-scale problem in skeleton extraction.

\begin{figure}[!t]
\centering
\includegraphics[width=1.0\linewidth]{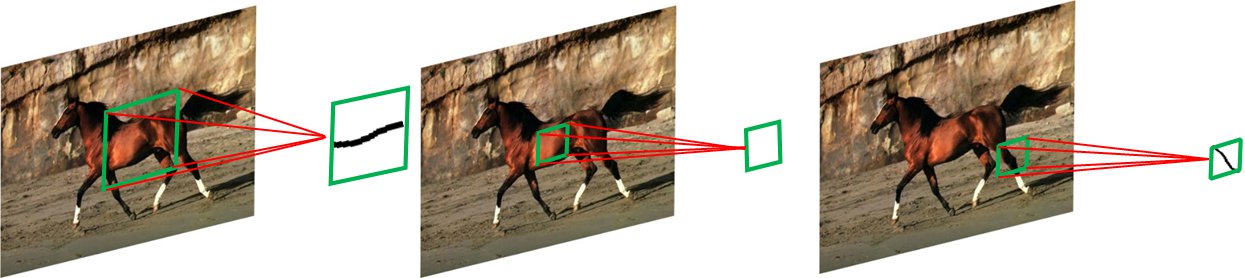}
\caption{Using filters (the green squares on images) of multiple sizes for skeleton extraction. Only when the size of the filter is larger than the scale of current skeleton part can the filter capture enough context feature to detect it.}\label{fig:resp}
\end{figure}

Referring to Fig.~\ref{fig:resp}, imagine that we are using multiple filters with different sizes (such as the convolutional kernels in convolutional networks) to detect a skeleton pixel at a specific scale; then only the filters with sizes larger than the scale will have responses, and others will not. Note that the sequential convolutional layers in a hierarchical network can be consider as filters with increasing sizes (the receptive field sizes of the original image of each convolutional layer are increasing from shallow to deep). So each convolutional layer is only able to capture the features of the skeleton pixels with scales less than its receptive field size. This sequence of increasing receptive field sizes provide a principle to quantize the skeleton scale space. With these observations, we propose to impose supervision at each side output (SO), optimizing them towards a scale-associated groundtruth skeleton map. More specifically, only skeleton pixels whose scales are smaller than the receptive field size of the SO are labeled by quantized scale values. The two sibling SOs at each stage are trained with multi-task loss for both skeleton localization and skeleton scale prediction. Thus the SOs at each stage are associated with specific scales and give a number of scale-specific skeleton score maps (the score map for one specified quantized scale value) as well as a skeleton scale map. Since the SOs in our network are scale-associated, we call them scale-associated side outputs (SSOs) and we refer to the SSOs for skeleton localization and skeleton scale prediction as Loc-SSO and ScalePred-SSO respectively.

The final predicted skeleton map is obtained by fusing Loc-SSOs. A straightforward fusion method is to average them. However, a skeleton pixel with large scale typically has a stronger response at the deeper SOs, and a weaker response at the shallower SOs; By contrast, a skeleton pixel with small scale may have strong responses at both of the two SOs. This motivates us to use a scale-specific weight layer to fuse the corresponding scale-specific skeleton score maps provided by each Loc-SSO.

In summary, the core contribution of this paper is the scale-associated side output layers, which enable both multi-task learning and fusion in a scale-depended way, to deal with the unknown scale problem. Therefore our network is able to detect skeleton pixels at multiple scales and estimate the scales.

To evaluate the performances of skeleton extraction methods, datasets with groundtruth skeleton maps as well as groudtruth scale maps are required. We constructed such a dataset in our previous work~\cite{Ref:ShenCVPR16}, which we called SK506\footnote{http://wei-shen.weebly.com/uploads/2/3/8/2/23825939/sk506.zip}. There are 506 natural images in this dataset, which were selected from the recent published MS COCO dataset~\cite{Ref:Chen15}. A skeletonization method~\cite{Ref:Bai07} was applied to the human-annotated foreground segmentation maps of the selected images to generate the groundtruth skeleton maps and the groundtruth scale maps. But the size of this dataset was small. Therefore, in this paper, we construct a larger dataset, containing $1,491$ natural images, annotated in the same way. We rename the SK506 dataset SK-SMALL and call the newly constructed one SK-LARGE. For consistency, SK-SMALL is a subset of SK-LARGE.

This paper extends our preliminary work~\cite{Ref:ShenCVPR16} by the following contributions: (1) Training the side outputs of each stage with a multi-task loss by introducing a new scale regression term. (2) Constructing a larger dataset for skeleton extraction. (3) More experimental results and discussions about the usefulness of the extracted skeletons in object detection applications.
\section{Related Works}
Object skeleton extraction has been studied a lot in recent decades. However, most works in the early stages~\cite{Ref:Saha15,Ref:Bai07} only focus on skeleton extraction from pre-segmented images. As these works make a strict assumption that object silhouettes are provided, i.e., the object has already been segmented, they cannot be applied to our task.

Pioneering researchers tried to extract skeletons from the gradient intensity maps computed on natural images. The gradient intensity map was typically obtained by applying directional derivative operators to a gray-scale image smoothed by a Gaussian kernel. For instance, in~\cite{Ref:Lindeberg98}, Lindeberg provided an automatic mechanism to determine the best size of the Gaussian kernel for gradient computation, and also proposed to detect skeletons as the pixels for which the gradient intensity takes a local maximum (minimum) in the direction of the main principal curvature. In~\cite{Ref:Lindeberg13}, he also gave a theoretic analysis of such scale selection mechanisms and showed that they are useful for other low level feature detection, such as interesting point detection. Majer~\cite{Ref:Majer04} pointed out that the second derivative of Gaussian filter kernel can detect skeletons under the assumption that skeletons are consider to be step or Gaussian ridge models. Jang and Hong~\cite{Ref:Jang01} extracted the skeleton from the pseudo-distance map which was obtained by iteratively minimizing an object function defined on the gradient intensity map. Yu and Bajaj~\cite{Ref:Yu04} proposed to trace the ridges of the skeleton intensity map calculated from the diffused vector field of the gradient intensity map, which can remove undesirablely biased skeletons. \cite{Ref:Liu98} was the pioneer for detecting symmetry and perform segmentation simultaneously by modeling and linking local symmetry parts, where skeleton extraction was formulated in terms of minimizing a goodness of fitness function defined on the gradient intensities. But due to the lack of supervised learning, these methods are only able to handle images with simple scenes.

Recent learning based skeleton extraction methods are better at dealing with complex scene. One type of methods formulates skeleton extraction as a per-pixel classification problem. Tsogkas and Kokkinos~\cite{Ref:Tsogkas12} computed hand-designed features of multi-scale and multi-orientation at each pixel, and employed multiple instance learning to determine whether it is symmetric\footnote{Although symmetry detection is not the same problem as skeleton extraction, we also compare the methods for it with ours, as skeletons can be considered a subset of symmetry.} or not. Shen \emph{et al.}~\cite{Ref:Shen16} then improved this method by training MIL models on automatically learned scale- and orientation-related subspaces. Sironi \emph{et al.}~\cite{Ref:SironiLF14} transformed the per-pixel classification problem to a regression one to achieve skeleton localization and learn the distance to the closest skeleton segment in scale-space. Another type of learning based methods aims to learn the similarity between local skeleton segments (represented by superpixel~\cite{Ref:Levinshtein09,Ref:Lee13} or spine model~\cite{Ref:Widynski14}), and links them by hierarchical clustering~\cite{Ref:Levinshtein09}, dynamic programming~\cite{Ref:Lee13} or particle filtering~\cite{Ref:Widynski14}. Due to the limited power of hand-designed features, these methods are not effective at detecting skeleton pixels with large scales, as large context information is needed.

Our method was inspired by~\cite{Ref:Xie15}, which developed a holistically-nested network for edge detection (HED). But detecting edges does not need to deal with scales explicitly. Using a local filter to detect an edge pixel, no matter what the size of the filter is, will give some response. So summing up the multi-scale detection responses, which occurs in the fusion layer in HED, is able to improve the performance of edge detection~\cite{Ref:Ren08,Ref:DollarZ15,Ref:Shen15}, while bringing false positives across the scales for skeleton extraction (see the results in Fig.~\ref{fig:inter}). There are three main differences between HED and our method. (1) We supervise the SOs of the network with different scale-associated groundtruths, but the groundtruths in HED are the same at all scales. (2) We use different scale-specific weight layers to fuse the corresponding scale-specific skeleton score maps provided by the SOs, while the SOs are fused by a single weight layer in HED. (3) We perform multi-task learning for the SOs of each stage by introducing a new scale regression loss, but only classification loss is considered in HED. The first two changes use the multi stages in a network to explicitly detect the unknown scale, which HED is unable to deal with. While the last change takes advantage of scale supervision to let our method provide a more informative result, i.e., the predicted scale for each skeleton pixel, which is useful for other potential applications, such as object segmentation and object proposal detection (we will show this in Sec.~\ref{sec:obj_seg} and Sec.~\ref{sec:detection}). By contrast, the output of HED cannot be applied to these applications.

\begin{figure}[!h]
\centering
\includegraphics[width=1.0\linewidth]{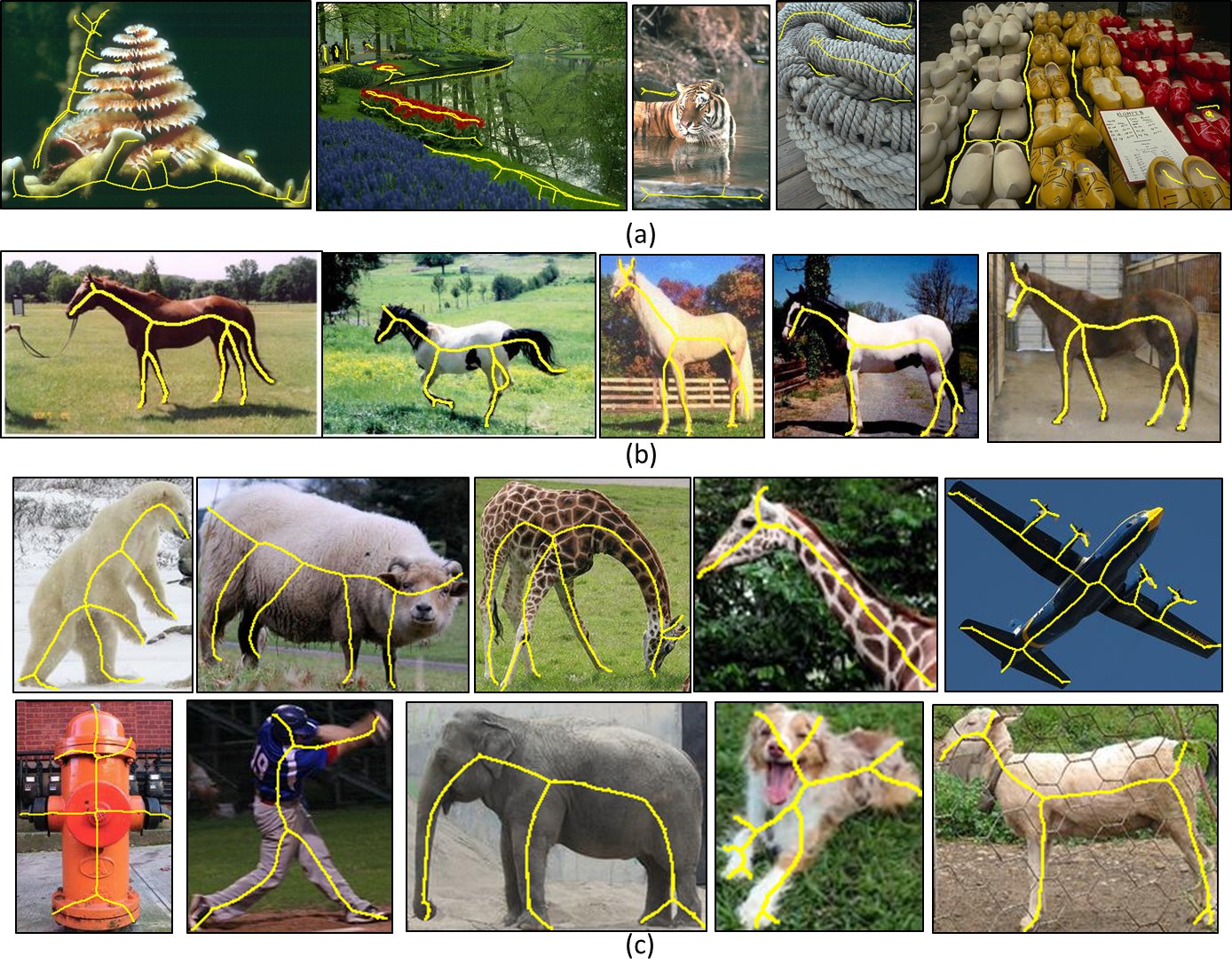}
\caption{Samples from three datasets. (a) The SYMMAX300 dataset~\cite{Ref:Tsogkas12}. (b) The WH-SYMMAX dataset~\cite{Ref:Shen16}. (c) SK-SMALL and SK-LARGE constructed by us. The groundtruths for skeletons and local reflection symmetries are in yellow.}\label{fig:dataset}
\end{figure}

There are only two other datasets related to our task. One is the SYMMAX300 dataset~\cite{Ref:Tsogkas12}, which is converted from the well-known Berkeley Segmentation Benchmark (BSDS300)~\cite{Ref:Martin01}. But this dataset is used mostly for local reflection symmetry detection. Local reflection symmetry~\cite{Ref:liu09,Ref:LeeL12} is a low-level feature of images, and does not depend on the concept of ``object''. Some examples from this dataset are shown in Fig.~\ref{fig:dataset}(a). Note that a large number of symmetries occur outside object. In general, the object skeletons are a subset of the local reflection symmetry. Another dataset is WH-SYMMAX~\cite{Ref:Shen16}, which is converted from the Weizmann Horse dataset~\cite{Ref:Borenstein02}. This dataset is suitable to verify object skeleton extraction methods; however, as shown in Fig.~\ref{fig:dataset}(b) a limitation is that only one object category, the horse, is contained in it. On the contrary, the objects, in our newly built dataset SK-LARGE, belong to a variety of categories, including humans, animals, such as birds, dogs and giraffes, and man made objects, such as planes and hydrants (Fig.~\ref{fig:dataset}(c)). Therefore, SK-LARGE not only contains more images, but also has more variability in object scales. We evaluate several skeleton extraction methods as well as symmetry detection methods on WH-SYMMAX, SK-SMALL and SK-LARGE. The experimental results demonstrate that our method significantly outperforms others.

\section{Methodology}
In this section, we describe our methods for object skeleton localization and scale prediction. First, we introduce the architecture of our network. Then, we discuss how to optimize and fuse the multiple scale-associated side outputs (SSOs) to extract the skeleton and predict the scale.
\subsection{Network Architecture}
\begin{figure*}[!th]
\centering
\includegraphics[width=0.8\linewidth]{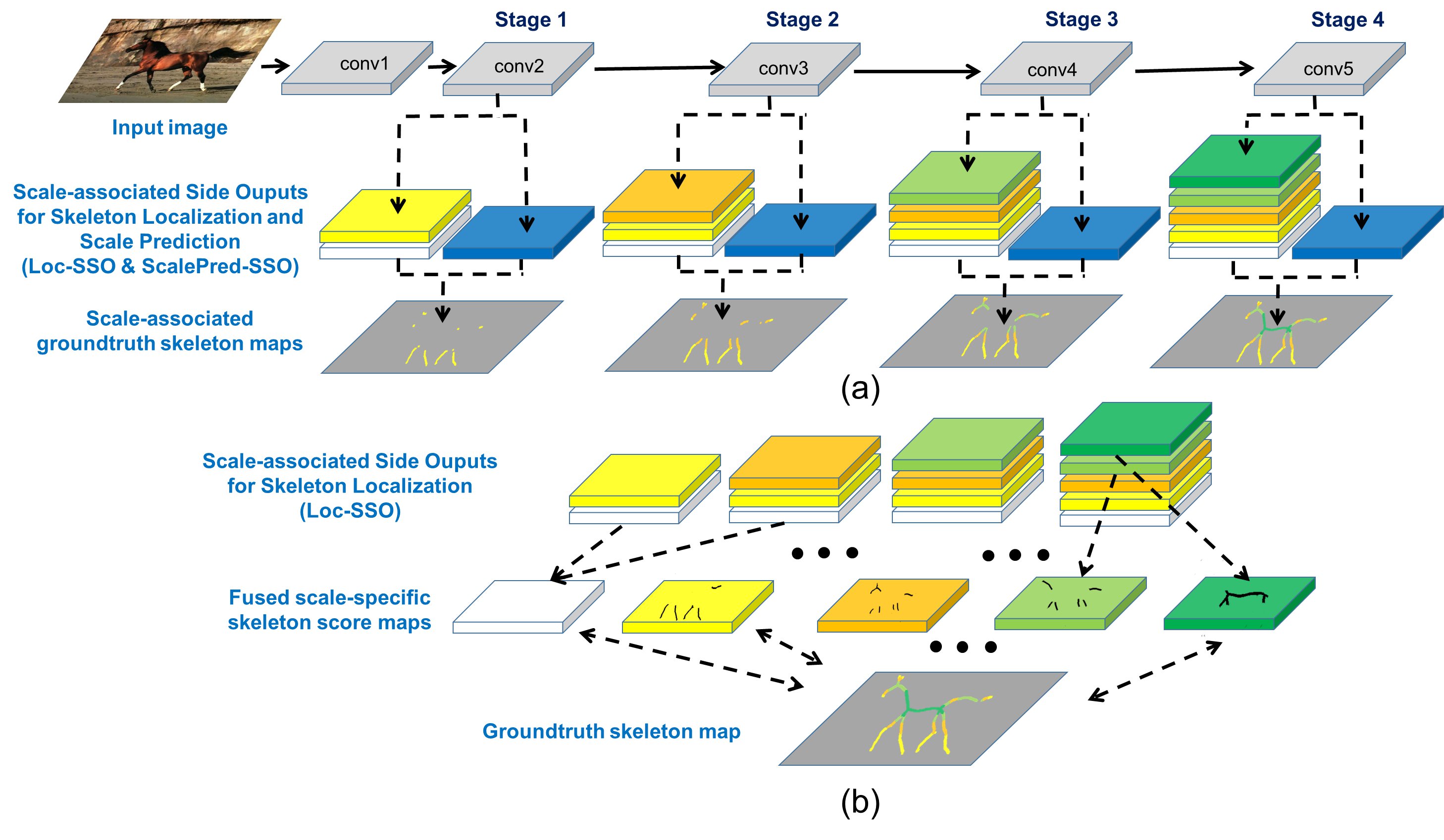}
\caption{The proposed network architecture for skeleton extraction, which is converted from VGG 16-layer net~\cite{Ref:Simonyan14}. (a) Multi-task Scale-associated side outputs (SSOs) learning. Our network has 4 stages with SSO layers connected to the convolutional layers. Each stage branches into two sibling SSO layers, one for skeleton localization and the other for scale prediction, denoted by Loc-SSO (the left multi-color blocks) and ScalePred-SSO (the right blue block), respectively. The SSOs in each stage are guided by a scale-associated groundtruth skeleton map (The skeleton pixels with different quantized scales are in different colors. Each block in a Loc-SSO is the activation map for one quantized scale, marked by the corresponding color). (b) Scale-specific fusion. Each Loc-SSO provides a certain number of scale-specific skeleton score maps (identified by stage number-quantized scale value pairs). The score maps of the same scales from different stages will be sliced and concatenated. Five scale-specific weighted-fusion layers are added to automatically fuse outputs from multiple stages.
}\label{fig:network}
\end{figure*}

We propose a new architecture for skeleton extraction, which is built on the HED network~\cite{Ref:Xie15}. HED is used for edge detection. Here, to address the unknown scale problem in skeleton extraction, we make two important modifications in our network: (a) we connect the proposed Loc-SSO and ScalePred-SSO layers to the last convolutional layer in each stage except for the first one, respectively conv2\_2, conv3\_3, conv4\_3, conv5\_3. The receptive field sizes of the sequential stages are 14, 40, 92, 196, respectively. The reason why we omit the first stage is that the receptive field size of the last convolutional layer is too small (only 5 pixels) to capture any skeleton features. There are only a few skeleton pixels with scales less than such a small receptive field. (b) Each Loc-SSO is connected to a slice layer to obtain the skeleton score map for each scale. Then from all these SO layers, we use a scale-specific weight layer to fuse the skeleton score maps for this scale. Such a scale-specific weight layer can be achieved by a convolutional layer with $1\times1$ kernel size. In this way, the skeleton score maps for different scales are fused by different weight layers. The fused skeleton score maps for each scale are concatenated together to form the final predicted skeleton map. An illustration for these two modifications are shown in Fig.~\ref{fig:network}(a) and Fig.~\ref{fig:network}(b), respectively. To sum up, our holistically-nested network architecture has 4 stages with additional SSO layers, with strides 2, 4, 8 and 16, respectively, and with different receptive field sizes; it also has 5 additional weight layers to fuse the Loc-SSOs.

\subsection{Skeleton Extraction by Learning Multi-task Scale-associated Side Outputs}
Skeleton localization can be formulated as a per-pixel classification problem. Given a raw input image $X=\{x_j,j=1,\ldots,|X|\}$, the goal is to predict its skeleton map $\hat{Y}=\{\hat{y}_j,j=1,\ldots,|X|\}$, where $\hat{y}_j\in\{0,1\}$ denotes the predicted label for each pixel $x_j$, i.e., if $x_j$ is predicted as a skeleton pixel, $\hat{y}_j=1$; otherwise, $\hat{y}_j=0$. Here, we also aim to predict the scale map $\hat{S}=\{\hat{s}_j,j=1,\ldots,|X|\}$, where $\hat{s}_j\in\mathbb{R}$, and $\hat{s}_j > 0$ if $\hat{y}_j=1$; otherwise $\hat{s}_j = 0$ if $\hat{y}_j=0$. This is a per-pixel regression problem. To sum up, our purpose is to address two tasks: One is skeleton localization, which takes input $X$ and outputs $\hat{Y}$; the other is scale prediction, whose input is $X$ and outputs $\hat{Y}$ and $\hat{S}$ simultaneously. By addressing the latter task, not only can the performance of the former be improved (Sec.~\ref{sec:des_eva}), but the object segmentation map can be obtained directly (Sec.~\ref{sec:obj_seg}). Next, we describe how to learn and fuse the SSOs in the training phase as well as how to use the learned network in the testing phase, respectively.
\subsubsection{Training Phase} \label{sec:training}
\begin{figure}[!th]
\centering
\includegraphics[width=0.8\linewidth]{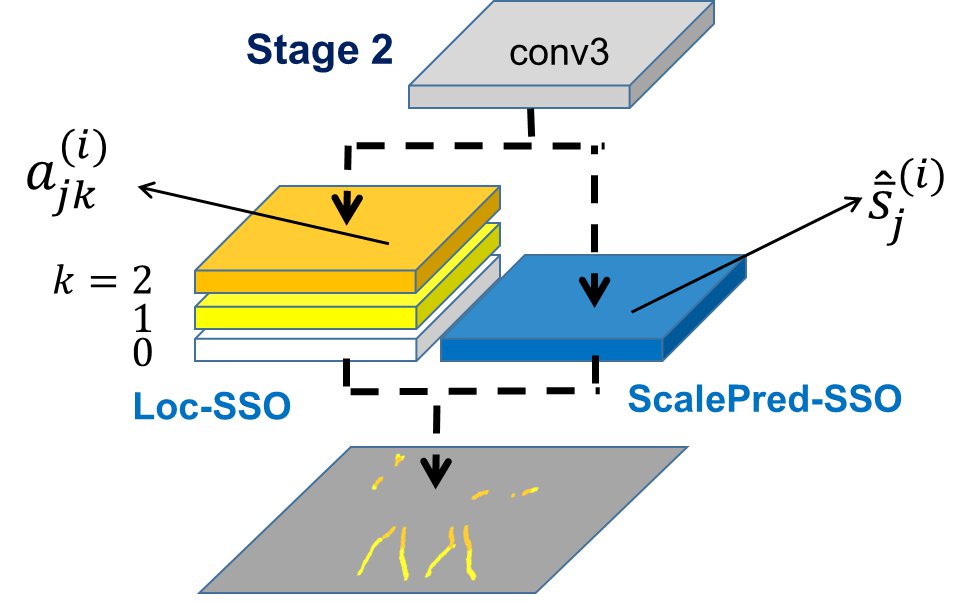}
\caption{An example of the computation of the scale-associated side outputs (SSOs) at each stage. The stage index is $2$. Thus, $i=2$, $K^{(i)}=2$. $a^{(i)}_{jk}$ and $\hat{\bar{s}}^{(i)}_j$ are the activations of the $i$-th Loc-SSO associated with the quantized scale $k$ and the $i$-th ScalePred-SSO for the input $x_j$, respectively. Please refer to text to see the meanings of the notations.
}\label{fig:notation}
\end{figure}

Following the definition of skeletons~\cite{Ref:Blum67}, we define the scale of each skeleton pixel as the diameter of the maximal disk centered at it, which can be obtained when computing the groundtruth skeleton map from the groundtruth segmentation map. So we are given a training dataset denoted by $\{(X^{(n)},Y^{(n)},S^{(n)}),n=1,\ldots,N\}$, where $X^{(n)}=\{x^{(n)}_j,j=1,\ldots,|X^{(n)}|\}$ is a raw input image and $Y^{(n)}=\{y^{(n)}_j,j=1,\ldots,|X^{(n)}|\}$ ($y^{(n)}_j\in\{0,1\}$) and $S^{(n)}=\{s^{(n)}_j,j=1,\ldots,|X^{(n)}|\}$ ($s^{(n)}_j\geq0$) are its corresponding groundtruth skeleton map and groundtruth scale map. Note that, we have $y^{(n)}_j=\mathbf{1}(s^{(n)}_j>0)$, where $\mathbf{1}(\cdot)$ is an indicator function. First, we describe how to compute a quantized skeleton scale map for each training image, which will be used for guiding the network training.
\paragraph{Skeleton scale quantization.}
As now we consider a single image, we drop the image superscript $n$. We aim to learn a network with multiple stages of convolutional layers linked with two sibling SSO layers. Assume that there are $M$ such stages in our network, in which the receptive field sizes of the convolutional layers increase in sequence. Let $(r_i;i=1,\ldots,M)$ be the sequence of the receptive field sizes. Recall that only when the receptive field size is larger than the scale of a skeleton pixel can the convolutional layer capture the features inside it. Thus, the scale of a skeleton pixel can be quantized into a discrete value, to indicate which stages in the network are able to detect this skeleton pixel. (Here, we assume that $r_M$ is sufficiently large to capture the features of the skeleton pixels with the maximum scale). The quantized value $z$ of a scale $s$ is computed by

\begin{equation}
z=\left\{
\begin{aligned}
&\arg\min_{i=1,\ldots,M} i, \; \text{s.t.} \; {r_i}>\rho{s}   &\text{if} \; s>0\\
&0 &\text{if} \; s=0 \\
\end{aligned}
\right.,
\end{equation}
where $\rho>1$ is a hyper parameter to ensure that the receptive field sizes are large enough for feature computation. (We set $\rho=1.2$ in our experiments.)
For an image $X$, we build a quantized scale value map $Z=\{z_j,j=1,\ldots,|X|\}\}$($z_j\in\{0,1,\ldots,M\}$).
\paragraph{Scale-associated side outputs learning for pixel classification.}
The groundtruth skeleton map $Y$ can be trivially computed from $Z$: $Y=\mathbf{1}(Z>0)$, but not vice versa. So we guide the network training by $Z$ instead of $Y$, since it gives more supervision. This converts a binary classification problem to a multi-class classification one, where each class corresponds to a quantized scale. Towards this end, each Loc-SSO layer in our network is associated with a softmax classifier. But according to the above discussions, each stage in our network is only able to detect the skeleton pixels at scales less than its corresponding receptive field size. Therefore, the side output is scale-associated. For the $i$-th Loc-SSO, we supervise it to a scale-associated groundtruth skeleton map: $Z^{(i)}=Z\circ{\mathbf{1}(Z\leq{i})}$, where $\circ$ is an element-wise product operator. Let $K^{(i)}=i$, then we have $Z^{(i)}=\{z^{(i)}_j,j=1,\ldots,|X|\},z^{(i)}_j\in\{0,1,\ldots,K^{(i)}\}$. To better understand this computation, we show an example of computing these variables in Fig.~\ref{fig:notation}.
Let $\ell_{cls}^{(i)}(\mathbf{W},\mathbf{\Phi}^{(i)})$ denote the loss function for this Loc-SSO, where $\mathbf{W}$ and $\mathbf{\Phi}^{(i)}$ are the layer parameters of the network and the parameters of the classifier of this stage. The loss function of our network is computed over all pixels in the training image $X$ and the scale-associated groundtruth skeleton map $Z^{(i)}$. Generally, the numbers of skeleton pixels at different scales are different and are much less than the number of non-skeleton pixels in an image. Therefore, we define a weighted softmax loss function to balance the loss between these multiple classes:
\begin{equation}
\begin{aligned}
&\ell_{cls}^{(i)}(\mathbf{W},\mathbf{\Phi}^{(i)})=  \\
&-\frac{1}{|X|}\sum_{j=1}^{|X|}\sum_{k=0}^{K^{(i)}}\beta_k^{(i)}\mathbf{1}(z^{(i)}_j=k)\log\text{Pr}(z^{(i)}_j=k|X;\mathbf{W},\mathbf{\Phi}^{(i)}),
\end{aligned}
\end{equation}
where $\beta_k^{(i)}$ is the loss weight for the $k$-th class and $\text{Pr}(z^{(i)}_j=k|X;\mathbf{W},\mathbf{\Phi}^{(i)})\in[0,1]$ is the predicted score given by the classifier for how likely the quantized scale of $x_j$ is $k$. Let $\mathcal {N}(\cdot)$ denote the number of non-zero elements in a set, then $\beta_k$ can be computed by
\begin{equation} \label{eqn:beta}
\beta_k^{(i)}=\frac{\frac{1}{\mathcal {N}(\mathbf{1}(Z^{(i)}==k))}}{\sum_{k=0}^{K^{(i)}}\frac{1}{\mathcal {N}(\mathbf{1}(Z^{(i)}==k))}}.
\end{equation}
Let $a^{(i)}_{jk}$ be the activation of the $i$-th Loc-SSO associated with the quantized scale $k$ for the input $x_j$ (Fig.~\ref{fig:notation}), then we use the softmax function~\cite{Ref:Bishop06} $\sigma(\cdot)$ to compute
\begin{equation}
\text{Pr}(z^{(i)}_j=k|X;\mathbf{W},\mathbf{\Phi}^{(i)})=\sigma(a^{(i)}_{jk})=\frac{\exp(a^{(i)}_{jk})}{\sum_{k=0}^{K^{(i)}}\exp(a^{(i)}_{jk})}.
\end{equation}
The partial derivation of $\ell_{cls}^{(i)}(\mathbf{W},\mathbf{\Phi}^{(i)})$ w.r.t. $a^{(i)}_{jl}$ ($l\in\{0,1,\ldots,K^{(i)}\}$) is obtained by
\begin{equation}
\begin{aligned}
&\frac{\partial{\ell_{cls}^{(i)}(\mathbf{W},\mathbf{\Phi}^{(i)})}}{\partial{a^{(i)}_{jl}}}=-\frac{1}{|X|}\bigg(\beta_{l}^{(i)}\mathbf{1}(z^{(i)}_j=l)-\\
&\sum_{k=0}^{K^{(i)}}\beta_{k}^{(i)}\mathbf{1}(z^{(i)}_j=k)\text{Pr}(z^{(i)}_j=l|X;\mathbf{W},\mathbf{\Phi}^{(i)})\bigg).
\end{aligned}
\end{equation}
\paragraph{Scale-associated side outputs learning for scale prediction.}
As we described, scale prediction is a per-pixel regression problem. In a regression problem, regression target normalization is a crucial pre-process. The receptive field size of each stage can serve as a good reference for scale normalization. For the $i$-th ScalePred-SSO, we guide it to a normalized scale-associated groundtruth skeleton map $\bar{S}^{(i)}=2\frac{Z^{(i)}{\circ}S}{r_i}-1$. This normalization maps each element $s_j$ in $S$ into the range $[-1, 1)$. Let $\hat{\bar{s}}^{(i)}_j$ be the predicted scale by the $i$-th ScalePred-SSO, i.e., the activation of the $i$-th ScalePred-SSO for the input $x_j$ (Fig.~\ref{fig:notation}), the regression loss is defined by
\begin{equation}
\ell_{reg}^{(i)}(\mathbf{W},\mathbf{\Psi}^{(i)})= \frac{\sum_{j=1}^{|X|}\mathbf{1}(z^{(i)}_j>0)\|\hat{\bar{s}}^{(i)}_j-\bar{s}^{(i)}_j\|_2^2}{\mathcal {N}(\mathbf{1}(Z^{(i)}>0))},
\end{equation}
where $\mathbf{\Psi}^{(i)}$ is the parameter of the regressor for $i$-th stage. Note that, for non skeleton pixels and those which have too large scale to be captured by this stage, do not contribute to the regression loss $\ell_{reg}^{(i)}$.
\paragraph{Multi-task loss.}
Each stage in our network has two sibling side output layers, i.e., Loc-SSO and ScalePred-SSO. We use a multi-task loss to jointly train them:
\begin{equation} \label{eqn:multi_task}
\ell_{s}^{(i)}(\mathbf{W},\mathbf{\Phi}^{(i)},\mathbf{\Psi}^{(i)}) = \ell_{cls}^{(i)}(\mathbf{W},\mathbf{\Phi}^{(i)}) + \lambda\ell_{reg}^{(i)}(\mathbf{W},\mathbf{\Psi}^{(i)}),
\end{equation}
where the hyper-parameter $\lambda$ controls the balance between the two task losses. Then the loss function for all the side outputs is simply obtained by
\begin{equation}
\mathcal {L}_s(\mathbf{W},\mathbf{\Phi},\mathbf{\Psi})=\sum_{i=1}^M\ell_{s}^{(i)}(\mathbf{W},\mathbf{\Phi}^{(i)},\mathbf{\Psi}^{(i)}).
\end{equation}
where $\mathbf{\Phi}=(\mathbf{\Phi}^{(i)};i=1,\ldots,M)$ and $\mathbf{\Psi}=(\mathbf{\psi}^{(i)};i=1,\ldots,M)$ denote the parameters of the classifiers and the regressors in all the stages, respectively.
\paragraph{Multiple scale-associated side outputs fusion.}
For an input pixel $x_j$, each scale-associated side output provides a predicted score $\text{Pr}(z^{(i)}_j=k|X;\mathbf{W},\mathbf{\Phi}^{(i)})$ (if $k{\leq}K^{(i)}$) for representing how likely its quantized scale is $k$. We can obtain a fused score $f_{jk}$ by simply summing them with weights $\mathbf{h}_k=({h}^{(i)}_k;i=\max(k,1),\ldots,M)$:
\begin{equation}
\begin{aligned}
f_{jk}=\sum^M_{i=\max(k,1)}{h}^{(i)}_k\text{Pr}(z^{(i)}_j=k|X;\mathbf{W},\mathbf{\Phi}^{(i)}),\\\text{s.t.}\;\sum_{i=\max(k,1)}^M{h}^{(i)}_k=1.
\end{aligned}
\end{equation}
We can understand the above fusion by this intuition: each scale-associated side output provides a certain number of scale-specific predicted skeleton score maps, and we use $M+1$ scale-specific weight layers: $\mathbf{H}=(\mathbf{h}_k;k=0,\ldots,M)$ to fuse them. Similarly, we can define a fusion loss function by
\begin{equation}
\begin{aligned}
&\mathcal {L}_f(\mathbf{W},\mathbf{\Phi},\mathbf{H})=  \\
&-\frac{1}{|X|}\sum_{j=1}^{|X|}\sum_{k=0}^{M}\beta_k\mathbf{1}(z_j=k)\log\text{Pr}(z_j=k|X;\mathbf{W},\mathbf{\Phi},\mathbf{h}_k),
\end{aligned}
\end{equation}
where $\beta_k$ is defined by the same way in Eqn.~\ref{eqn:beta} and $\text{Pr}(z_j=k|X;\mathbf{W},\mathbf{\Phi},\mathbf{w}_k)=\sigma(f_{jk})$.

Finally, we can obtain the optimal parameters by
\begin{equation}
(\mathbf{W},\mathbf{\Phi},\mathbf{\Psi},\mathbf{H})\ast =\arg\min(\mathcal {L}_s(\mathbf{W},\mathbf{\Phi},\mathbf{\Psi})+\mathcal {L}_f(\mathbf{W},\mathbf{\Phi},\mathbf{H})).
\end{equation}
\subsubsection{Testing Phase}
Given a testing image $X=\{x_j,j=1,\ldots,|X|\}$, with the learned network $(\mathbf{W},\mathbf{\Phi},\mathbf{\Psi},\mathbf{H})\ast$, its predicted skeleton map $\hat{Y}=\{\hat{y}_j,j=1,\ldots,|X|\}$ is obtained by
\begin{equation} \label{eqn:predict}
\hat{y}_j = 1 - \text{Pr}(z_j=0|X;\mathbf{W}\ast,\mathbf{\Phi}\ast,\mathbf{h}_0\ast).
\end{equation}
Recall that $z_j=0$ and $z_j>0$ mean that $x_j$ is a non-skeleton/skeleton pixel, respectively. To predict the scale for each $x_j$, we first find its most likely quantized scale by
\begin{equation}
i\ast=\arg\max_{i=(1,\ldots,M)}\text{Pr}(z_j=i|X;\mathbf{W}\ast,\mathbf{\Phi}\ast,\mathbf{h}_i\ast).
\end{equation}
Then the predicted scale $\hat{s}_j$ is computed by
\begin{equation}
\hat{s}_j=\frac{\hat{\bar{s}}^{(i\ast)}_j + 1}{2}r_{i\ast},
\end{equation}
where $\hat{\bar{s}}^{(i\ast)}_j$ is the activation of the $i\ast$-th ScalePred-SSO.
We refer to our method as LMSDS, for learning multi-task scale-associated deep side outputs.
\subsection{Understanding of the Proposed Method}
\begin{figure}[!h]
\centering
\includegraphics[width=1.0\linewidth]{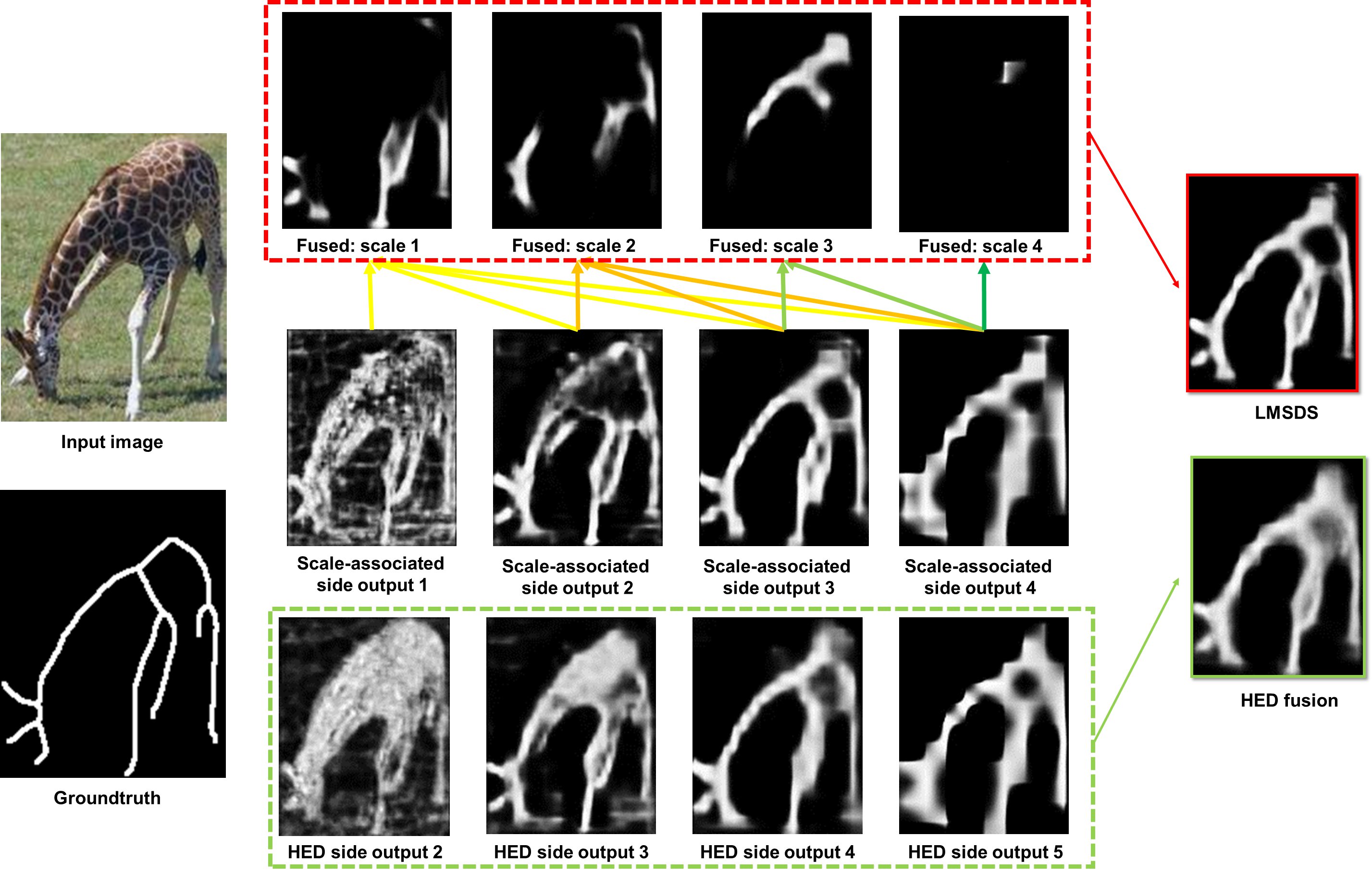}
\caption{The comparison between the intermediate results of LMSDS and HED. We observe that the middle row are able to differentiate skeleton pixels with different scales, while the latter cannot.
}\label{fig:inter}
\end{figure}

To understand our method more deeply, we illustrate the intermediate results and compare them with those of HED in Fig.~\ref{fig:inter}. The response of each Loc-SSO can be obtained by the similar way of Eqn.~\ref{eqn:predict}. We compare the response of each Loc-SSO to the corresponding side output in HED (The side output 1 in HED is connected to conv1\_2, while ours start from conv2\_2.). With the extra scale-associated supervision, the responses of our side outputs are indeed related to scale. For example, the first side output fires on the structures with small scales, such as the legs, the interior textures and the object boundaries; while in the second one, the skeleton parts of the head and neck become clear and meanwhile the noises on small scale structure are suppressed. In addition, we perform scale-specific fusion, by which each fused scale-specific skeleton score map corresponds to one scale, e.g., the first three response maps in Fig.~\ref{fig:inter} corresponding to legs, neck and torso respectively. By contrast, the side outputs in HED are not able to differentiate skeleton pixels with different scales. Consequently, the first two respond on the whole body, which causes false positives to the final fusion one.
\section{Experimental Results}
In this section we discuss the implementation details and compare the performance of our skeleton extraction methods with competitors.
\subsection{Implementation Details}
Our implementation is based on ``Caffe''~\cite{Ref:JiaCaffe14} and our architecture is built on the public available implementation of FCN~\cite{Ref:LongSD15} and HED~\cite{Ref:Xie15}. The whole network is fine-tuned from an initialization with the pre-trained VGG 16-layer net~\cite{Ref:Simonyan14}.This net is pre-trained on the subset of ImageNet used in an image classification challenge, called ILSVRC-2014~\cite{Ref:Russakovsky14}, which has 1000 categories and 1.2 million images.
\paragraph{Groundtruth generation}
The groundtruth skeleton map for each image is computed from its corresponding human-annotated foreground segmentation mask (1 for foreground objects and 0 for background). We apply a binary image skeletonization method based on the distance transform~\cite{Ref:Bai07} to these segmentation masks to generate the skeleton maps (1 for skeleton pixels and 0 for non-skeleton pixels) and use them as the groundtruths. The groundtruth scale of each skeleton pixel is two times of the minimal distance between this skeleton pixel and the boundary of the corresponding foreground segmentation mask.
\paragraph{Model parameters} The hyper parameters of our network include: mini-batch size (1), base learning rate ($1\times10^{-6}$), loss weight for each side-output (1), momentum (0.9), initialization of the nested filters(0), initialization of the scale-specific weighted fusion layer ($1/n$, where $n$ is the number of sliced scale-specific maps), the learning rate of the scale-specific weighted fusion layer ($5\times10^{-6}$), weight decay ($2\times10^{-4}$), maximum number of training iterations ($20,000$).
\paragraph{Data augmentation}
Data augmentation is a standard way to generate sufficient training data for learning a ``good'' deep network. We rotate the images to 4 different angles ($0^\circ$, $90^\circ$, $180^\circ$, $270^\circ$) and flip them with different axis (up-down, left-right, no flip), then resize images to 3 different scales ($0.8$, $1.0$, $1.2$), totally leading to an augmentation factor of 36. Note that when resizing a groundtruth skeleton map, the scales of the skeleton pixels in it should be multiplied by a resize factor accordingly.
\subsection{Skeleton Localization}
\subsubsection{Evaluation Protocol}
To evaluate skeleton localization performances, we follow the protocol used in~\cite{Ref:Tsogkas12}, under which the detected skeletons are measured by their maximum \emph{F-measure} ($\frac{2\cdot\text{Precision}\cdot\text{Recall}}{\text{Precision}+\text{Recall}}$) as well as precision-recall curves with respect to the groundtruth skeleton map. To obtain the precision-recall curves, the detected skeleton response is first thresholded into a binary map, which is then matched with the groundtruth skeleton map. The matching allows small localization errors between detected positives and groundtruths. If a detected positive is matched with at least one groundtruth skeleton pixel, it is classified as a true positive. By contrast, pixels that do not correspond to any groundtruth skeleton pixel are false positives. By assigning different thresholds to the detected skeleton response, we obtain a sequence of precision and recall pairs, which is used
to plot the precision-recall curve.
\subsubsection{Design Evaluation} \label{sec:des_eva}
The main difference between LMSDS and our preliminary work~\cite{Ref:ShenCVPR16}, FSDS, is that we apply multi-task learning in LMSDS.  Since the two tasks influence
each other through their shared representation (convolutional features), we can ask how multi-task learning influences the result of skeleton localization?

To answer this question, we compare the skeleton localization performances of these two methods on three datasets: SK-LARGE, SK-SMALL and WH-SYMMAX. Note that, by setting $\lambda=0$ in Eqn.~\ref{eqn:multi_task}, LMSDS reduces to FSDS. The comparison is summarized in Table~\ref{tbl:multi-task}, from which we observe that training with multi-task loss leads to a slight decrease in skeleton localization performance on SK-SMALL, but yeilds considerable improvements on SK-LARGE and WH-SYMMAX. The reason why the results are opposite on SK-SMALL and SK-LARGE may be because scale prediction is more difficult than skeleton localization, i.e., training a good model by using multi-task loss requires more training data. Although the training set of WH-SYMMAX is small, the variance of the data is also small, because only one object category is contained in it. To sum up, we argue that multi-task
training with sufficient training data can improve pure skeleton localization compared to training for skeleton localization alone. In Sec.~\ref{sec:obj_seg}, we will show that multi-task learning is important to obtain accurate predicted scales, which is useful for skeleton based object segmentation.

\begin{figure}[!h]
\centering
\includegraphics[trim=1cm 0cm 1cm 1cm, clip=true, width=1.0\linewidth]{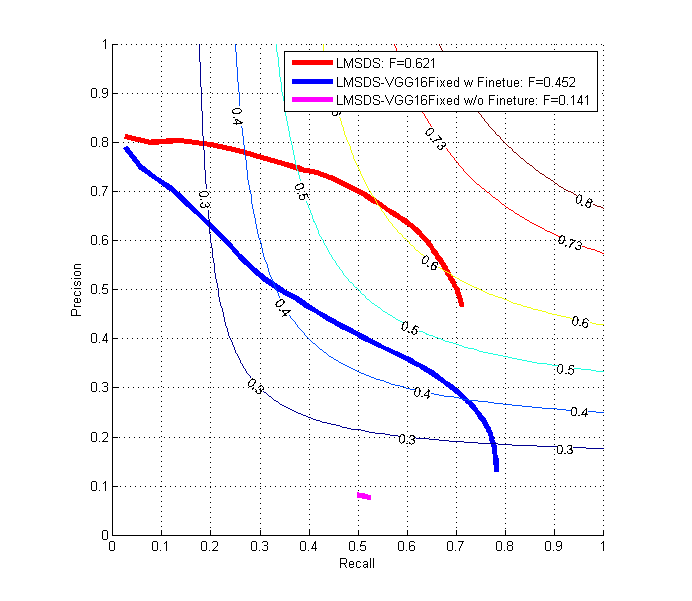}
\caption{The comparison between different network parameter settings in LMSDS.
}\label{fig:lmsds_vs_vgg16}
\end{figure}

\begin{table}[!h]
\centering
\caption{The validation of the influence of multi-task training on skeleton localization. The localization results are measured by their F-meansures.}\label{tbl:multi-task}
\begin{tabular}{cccc}
\toprule
&SK-SMALL&SK-LARGE&WH-SYMMAX\\
\midrule
FSDS&\textbf{0.623}&0.633&0.769\\
LMSDS&0.621&\textbf{0.649}&\textbf{0.779}\\
\bottomrule
\end{tabular}
\end{table}
Since our network is finetuned from the pre-trained VGG 16-layer net, another question is does the pre-trained VGG 16-layer net already have the ability to detect skeletons? To verify this, we consider two network parameter settings. One is we fix the weights of the VGG part in our network and train the rest part (denoted by LMSDS-VGGFixed w Finetune), the other is we fix the weights of the VGG part in our network and leave the rest in random initialization (denoted by LMSDS-VGGFixed w/o Finetune). As shown in Fig.~\ref{fig:lmsds_vs_vgg16}, the performance of ``LMSDS-VGGFixed w Finetune'' drops significantly and ``LMSDS-VGGFixed w/o Finetune'' even does not work (The skeleton detection results are nearly random noises. So for all the points on its precision-recall curve, the precision is very low and the recall is near $0.5$.). This result demonstrates that the pre-trained VGG 16-layer net is purely for the initialization of a part of our network, e.g., it does not
initialize the weights for the SSOs layers, and final weights of our network differ enormously from the initial weights. Consequently, the pre-trained VGG 16-layer net does not have the ability to detect skeletons.
\subsubsection{Performance Comparison}
\begin{figure}[!h]
\centering
\includegraphics[trim=0.5cm 0.5cm 1.5cm 0.5cm, clip=true, width=0.8\linewidth]{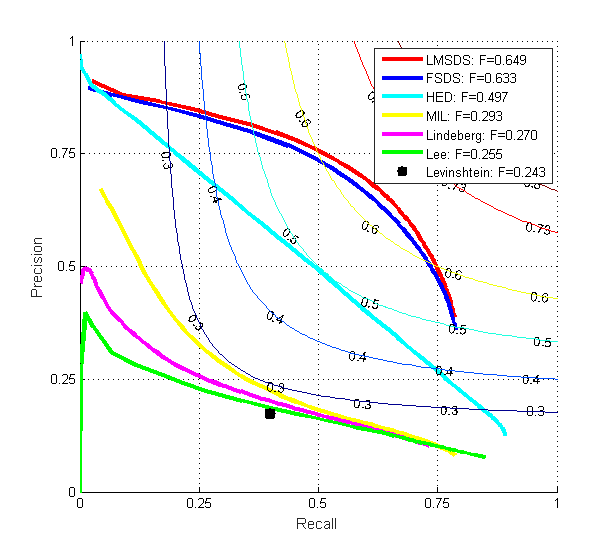}
\caption{Skeleton localization evaluation of skeletons extracted on SK-LARGE, which consists of 746 training and 745 testing images. Leading skeleton extraction methods are ranked according to their best F-measure with respect to groundtruth skeletons. LMSDS and FSDS achieve the top and the second best results, respectively. See Table~\ref{tbl:sk_large} for more details about the other quantity (Avg Runtime) and citations to competitors.\label{fig:pr_sk_large}
}\end{figure}

\begin{figure*}[!th]
\centering
\includegraphics[trim=0.4cm 0cm 0cm 0cm, clip=true, width=0.8\linewidth]{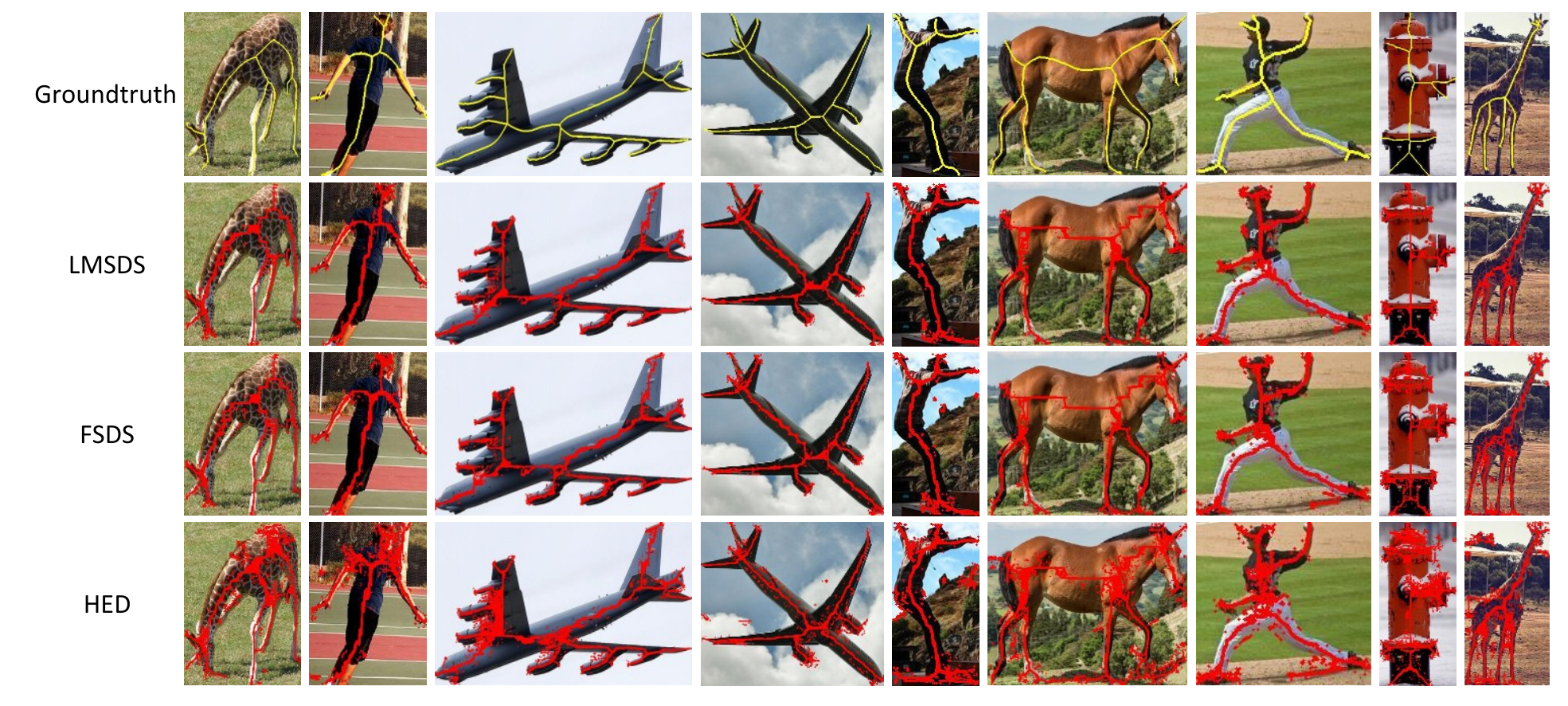}
\caption{Illustration of skeleton extraction results on SK-LARGE for several selected images. The groundtruth skeletons are in yellow and the thresholded extraction results are in red. Thresholds were optimized over the whole dataset.\label{fig:examples_sk}
}\end{figure*}

We conduct our experiments by comparing our method LMSDS with others, including a traditional image processing method (Lindeberg's method~\cite{Ref:Lindeberg98}), three learning based segment linking methods ( Levinshtein's method~\cite{Ref:Levinshtein09}, Lee's method~\cite{Ref:Lee13} and Particle Filter~\cite{Ref:Widynski14}), three per-pixel classification/regression methods (Distance Regression~\cite{Ref:SironiLF14}, MIL~\cite{Ref:Tsogkas12} and MISL~\cite{Ref:Shen16}) and two deep learning based method (HED~\cite{Ref:Xie15} and FSDS~\cite{Ref:ShenCVPR16}). For all theses methods, we use the source code provided by the authors with the default setting. For HED, FSDS and LMSDS, we perform sufficient iterations to ensure convergence. We apply a standard non-maximal suppression algorithm~\cite{Ref:DollarZ15} to the response maps of HED and ours to obtain the thinned skeletons for performance evaluation.
\paragraph{SK-LARGE}
We first conduct our experiments on our newly built SK-LARGE dataset. Object skeletons in this dataset have large variabilities in both structures and scales. We split this dataset into 746 training and 745 testing images. We report the F-measure as well as the average runtime per image of each method on this dataset in Table.~\ref{tbl:sk_large}. Observed that, both traditional image processing and per-pixel/segment learning methods do not perform well, indicating the difficulty of this task. Moreover, the segment linking methods are extremely time consuming. Our method LMSDS outperforms others significantly, even compared with the deep learning based method HED. In addition, thanks to the powerful convolution computation ability of GPU, our method can process images in real time, about 20 images per second. The precision/recall curves shown in Fig.~\ref{fig:pr_sk_large} show again that LMSDS is better than the alternatives, as ours gives both improved recall and precision in most of the precision-recall regimes. We illustrate the skeleton extraction results obtained by several methods in Fig.~\ref{fig:examples_sk} for qualitative comparison. These qualitative examples show that our method detects more groundtruth skeleton points and also suppresses false positives. The false positives in the results of HED are probably introduced because it does not use learning to combine different scales. Benefiting from scale-associated learning and scale-specific fusion, our method is able to suppress these false positives.
\begin{table}[!h]
\centering
\caption{Skeleton localization performance comparison between different methods on SK-LARGE. $\dag$GPU time.}\label{tbl:sk_large}
\begin{tabular}{ccc}
\toprule
Method&F-measure&Avg Runtime (sec)\\
\midrule
Lindeberg~\cite{Ref:Lindeberg98}&0.270&4.05\\
Levinshtein~\cite{Ref:Levinshtein09}&0.243&146.21\\
Lee~\cite{Ref:Lee13}&0.255&609.10\\
MIL~\cite{Ref:Tsogkas12}&0.293&42.40\\
HED~\cite{Ref:Xie15}&0.497&0.05$\dag$\\
FSDS (ours)&0.633&0.05$\dag$\\
LMSDS (ours)&0.649&0.05$\dag$\\
\bottomrule
\end{tabular}
\end{table}

\begin{figure}[!h]
\centering
\includegraphics[trim=1.0cm 1.0cm 1.5cm 1.0cm, clip=true, width=0.8\linewidth]{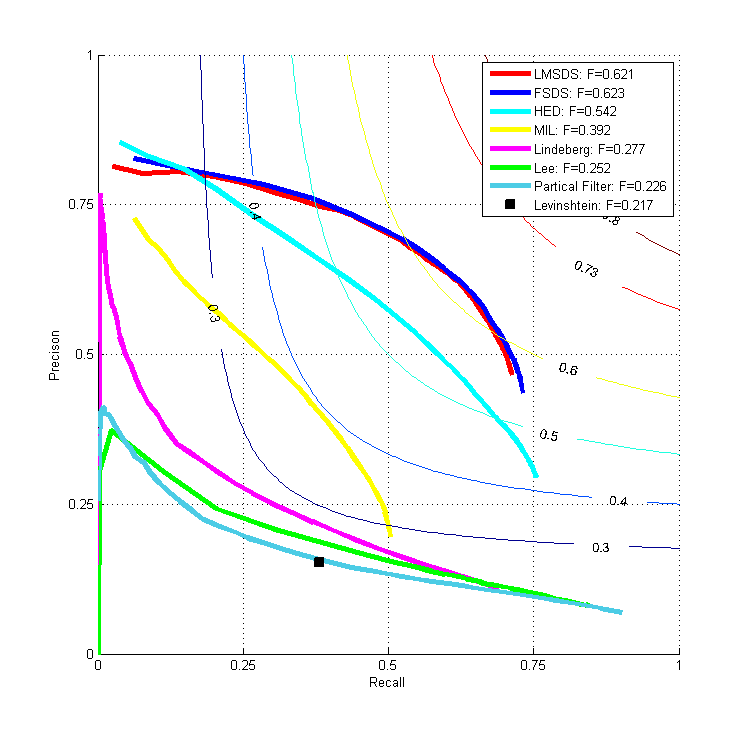}
\caption{Skeleton localization evaluation of skeleton extractors on SK-SMALL, which consists of 300 training and 206 testing images. Skeleton extraction methods are measured by their best F-measure with respect to groundtruth skeletons. FSDS and LMSDS achieve the top and the second best results, respectively. See Table~\ref{tbl:sk_small} for more details about the other quantity (Avg Runtime) and citations to competitors.\label{fig:pr_sk}
}\end{figure}

\paragraph{SK-SMALL}
We then perform comparisons on SK-SMALL. The training and testing sets of SK-SMALL contain $300$ and $206$ images, respectively. From the precision/recall curves shown in Fig.~\ref{fig:pr_sk} and summary statistics reported in Table.~\ref{tbl:sk_small}, we observe that LMSDS outperforms the others, except for our preliminary method, FSDS. LMSDS performs slightly worse on skeleton localization on SK-SMALL, for reasons we discussed in Sec.~\ref{sec:des_eva}.
\begin{table}[!h]
\centering
\caption{Skeleton localization performance comparison between different methods on SK-SMALL. $\dag$GPU time.}\label{tbl:sk_small}
\begin{tabular}{ccc}
\toprule
Method&F-measure&Avg Runtime (sec)\\
\midrule
Lindeberg~\cite{Ref:Lindeberg98}&0.277&4.03\\
Levinshtein~\cite{Ref:Levinshtein09}&0.218&144.77\\
Lee~\cite{Ref:Lee13}&0.252&606.30\\
Particle Filter~\cite{Ref:Widynski14}&0.226&322.25$\dag$\\
MIL~\cite{Ref:Tsogkas12}&0.392&42.38\\
HED~\cite{Ref:Xie15}&0.542&0.05$\dag$\\
FSDS (ours)&0.623&0.05$\dag$\\
LMSDS (ours)&0.621&0.05$\dag$\\
\bottomrule
\end{tabular}
\end{table}

\begin{figure}[!t]
\centering
\includegraphics[width=1.0\linewidth]{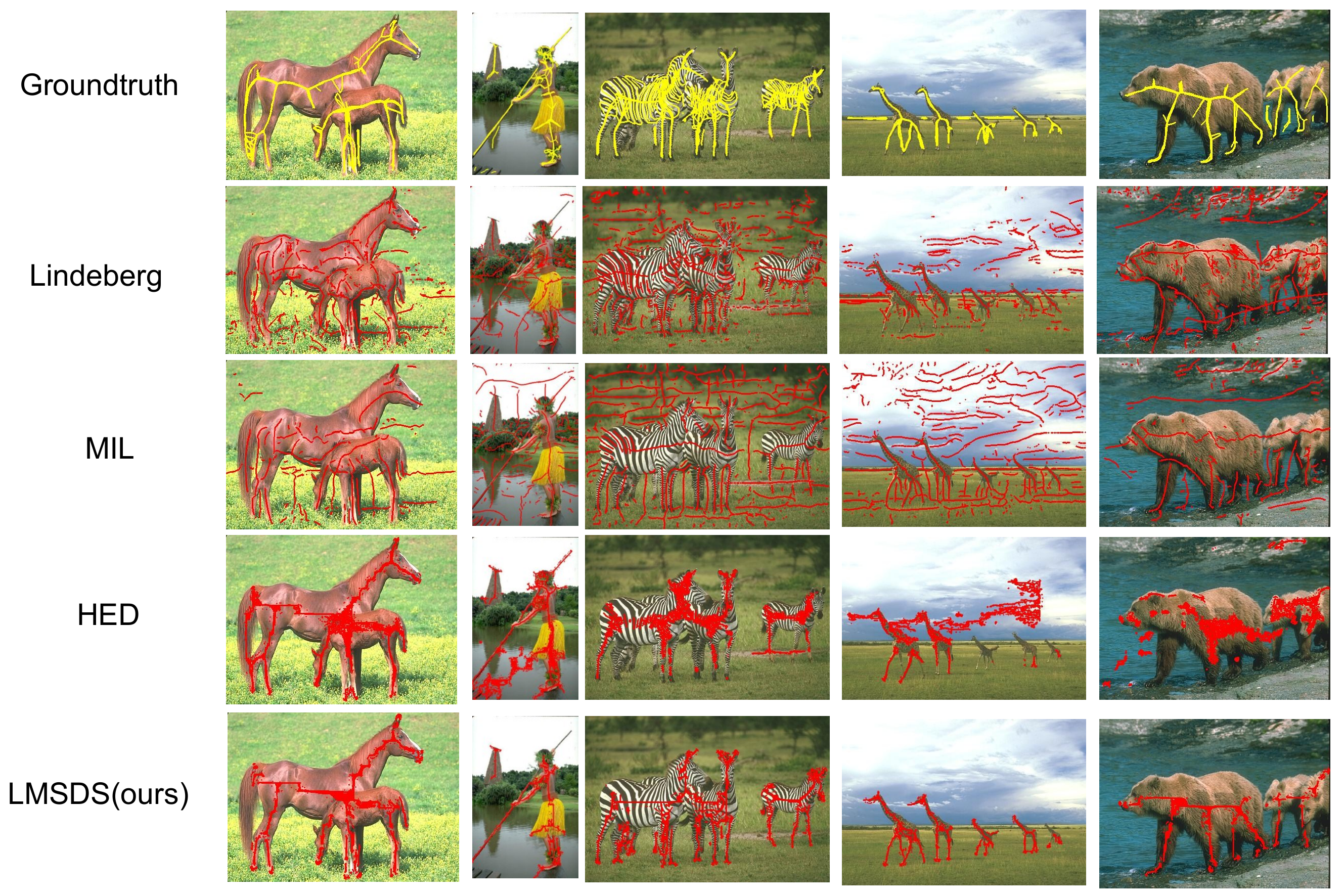}
\caption{Illustration of skeleton extraction results on the SYMMAX300 dataset~\cite{Ref:Tsogkas12} for several selected images. The groundtruth skeletons are in yellow and the thresholded extraction results are in red. Thresholds were optimized over the whole dataset.\label{fig:examples_sym}
}\end{figure}

\begin{figure}[!h]
\centering
\includegraphics[trim=0.5cm 1.0cm 1.5cm 0.5cm, clip=true, width=0.8\linewidth]{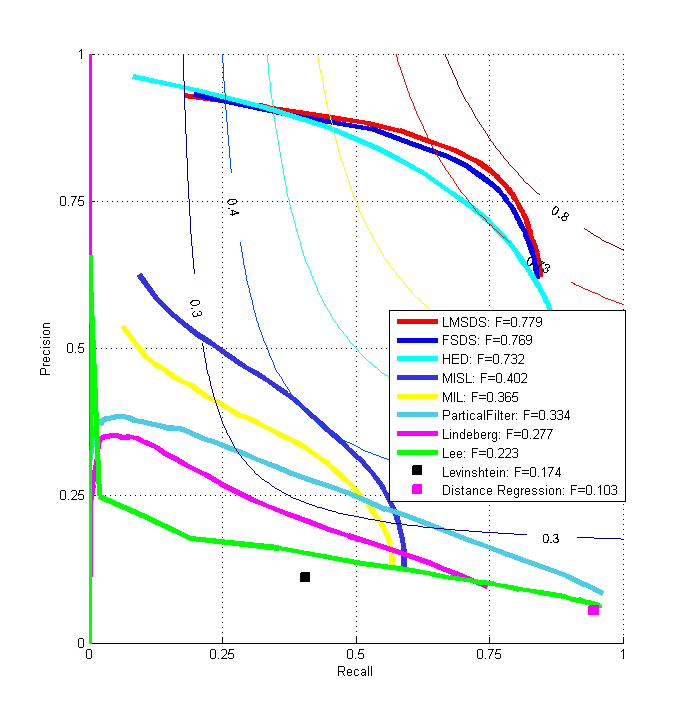}
\caption{Evaluation of skeleton extractors on WH-SYMMAX~\cite{Ref:Shen16}, which consists of 228 training and 100 testing images. Leading skeleton extraction methods are ranked according to their best F-measure with respect to groundtruth skeletons. Our method, FSDS achieves the top result and shows both improved recall and precision at most of the
precision-recall regime. See Table~\ref{tbl:horse} for more details about the other quantity (Avg Runtime) and citations to competitors.\label{fig:pr_horse}
}\end{figure}

\begin{figure*}[!th]
\centering
\includegraphics[width=0.8\linewidth]{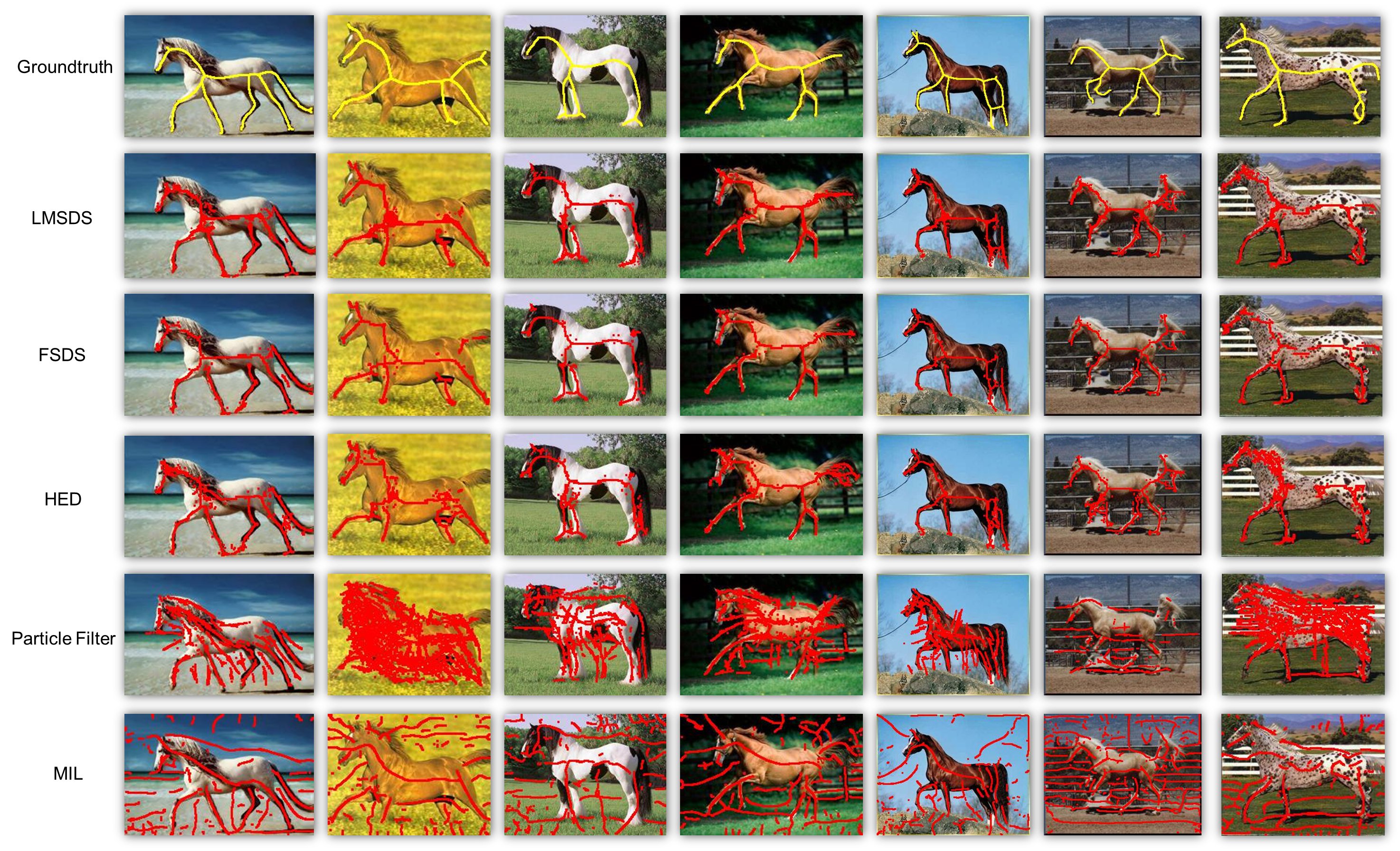}
\caption{Illustration of skeleton extraction results on WH-SYMMAX~\cite{Ref:Shen16} for several selected images. The groundtruth skeletons are in yellow and the thresholded extraction results are in red. Thresholds were optimized over the whole dataset.\label{fig:examples_horse}
}\end{figure*}

\paragraph{WH-SYMMAX}
The WH-SYMMAX dataset~\cite{Ref:Shen16} contains 328 images, of which the first 228 are used for training and the rest are used for testing. The precision/recall curves of skeleton extraction methods are shown in Fig.~\ref{fig:pr_horse} and summary statistics are in Table~\ref{tbl:horse}. Qualitative comparisons are
illustrated in Fig.~\ref{fig:examples_horse}. Both quantitative and qualitative results demonstrate that our method is clearly better than others.
\begin{table}[!h]
\centering
\caption{Skeleton localization performance comparison between different methods on WH-SYMMAX~\cite{Ref:Shen16}. $\dag$GPU time.}\label{tbl:horse}
\begin{tabular}{ccc}
\toprule
Method&F-measure&Avg Runtime (sec)\\
\midrule
Lindeberg~\cite{Ref:Lindeberg98}&0.277&5.75\\
Levinshtein~\cite{Ref:Levinshtein09}&0.174&105.51\\
Lee~\cite{Ref:Lee13}&0.223&716.18\\
Particle Filter~\cite{Ref:Widynski14}&0.334&13.9$\dag$\\
Distance Regression~\cite{Ref:SironiLF14}&0.103&5.78\\
MIL~\cite{Ref:Tsogkas12}&0.365&51.19\\
MISL~\cite{Ref:Shen16}&0.402&78.41\\
HED~\cite{Ref:Xie15}&0.732&0.06$\dag$\\
FSDS (ours)&0.769&0.07$\dag$\\
LMSDS (ours)&0.779&0.07$\dag$\\
\bottomrule
\end{tabular}
\end{table}
\paragraph{Skeleton Extraction for Multiple Objects}
Our method does not have the constraint that one image can only contain a single object. Here, we directly apply our model trained on SK-SMALL to images from SYMMAX300~\cite{Ref:Tsogkas12}, which contain multiple objects and complex background, e.g., the merged zebras. As the comparison shows in Fig.~\ref{fig:examples_sym}, our method can obtain
good skeletons for each object in these images, which have significantly less false positives corresponding to background and interior textures.
%
\paragraph{Cross Dataset Generalization}
A concern is that the scale-associated side outputs learned from one dataset might lead to higher generalization error when applied them to another dataset. To explore whether this is
the case, we test the model learned from one dataset on another one. For comparison, we list the cross dataset generalization results of MIL~\cite{Ref:Tsogkas12}, HED~\cite{Ref:Xie15} and our method in Table~\ref{tbl:cross}. Our method achieves better cross dataset generalization results than both the ``non-deep'' method (MIL) and the ``deep'' method (HED).

\begin{table}[!htbp]
\centering
\caption{Cross-dataset generalization results. TRAIN/TEST indicates the training/testing dataset used.}\label{tbl:cross}
\begin{tabular}{ccc}
\toprule
Method&Train/Test&F-measure\\
\midrule
MIL~\cite{Ref:Tsogkas12}&SK-LARGE/WH-SYMMAX&0.350\\
HED~\cite{Ref:Xie15}&SK-LARGE/WH-SYMMAX&0.583\\
LMSDS (ours)&SK-SMALL/WH-SYMMAX&0.701\\
\hline
MIL~\cite{Ref:Tsogkas12}&WH-SYMMAX/SK-LARGE&0.357\\
HED~\cite{Ref:Xie15}&WH-SYMMAX/SK-LARGE&0.420\\
LMSDS (ours)&WH-SYMMAX/SK-LARGE&0.474\\

\bottomrule
\end{tabular}
\end{table}

\subsection{Object Segmentation} \label{sec:obj_seg}
\begin{figure}[!t]
\centering
\includegraphics[width=0.8\linewidth]{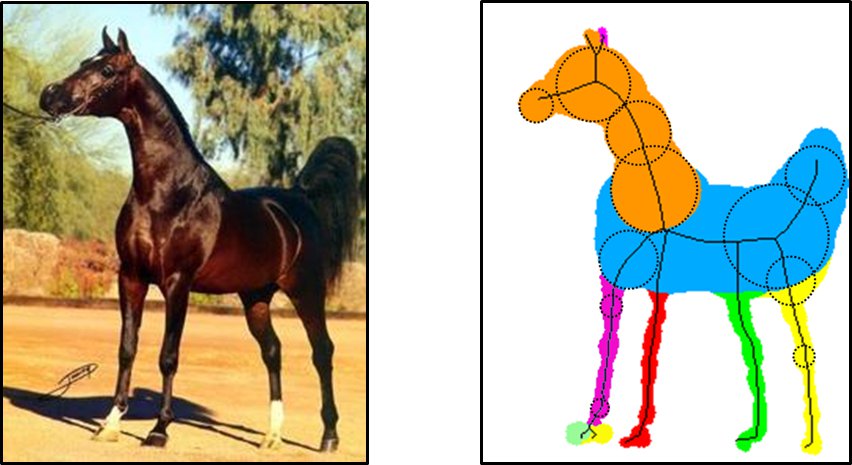}
\caption{Skeleton based object segmentation. Left: The original image. Right: The object segments reconstructed from the skeleton with scales. Different object segments are marked in different colors. The dashed circles are sampled maximal disks. \label{fig:sk_re}
}\end{figure}

We can use the predicted scale for each skeleton pixel to segment the foreground objects in images. For each skeleton pixel $x_j$, let $\hat{s}_j$ be its predicted scale, then for a skeleton segment $\{x_j,j=1,\ldots,N\}$, where $N$ is the number of the skeleton pixels in this segment, we obtain a object segment mask by $\mathcal {M}=\bigcup_{j=1}^ND_j$, where $D_j$ is the disk of center $x_j$ and diameter $\hat{s}_j$. Fig.~\ref{fig:sk_re} illustrates an example of object segments obtained by the above process. The more accurate the predicted scales are, the more better segmentation results. Therefore, evaluating the object segmentation results, not only can we validate the performance for skeleton extraction, but the potential usefulness of the obtained skeletons for high level vision tasks can be demostrated.
\subsubsection{Evaluation Protocol}
Following~\cite{Ref:Kim12,Ref:Carreira12,Ref:Alpert12}, we evaluate object segmentation results by assessing their consistency with the groundtruth object segmentation.
Two evaluation metrics are adopted here. One is the \emph{F-measure}~\cite{Ref:Alpert12}, which calculates the average best F-score between the groundtruth object segments and
the generated segments (for each groundtruth object segment, find the generated one with highest F-score, then these F-scores are averaged over the whole dataset).
The other is the \emph{Covering} metric~\cite{Ref:Kim12,Ref:Carreira12}, which calculates the average best overlapping score between groundtruth object segments and generated segments,
weighted by the object size. Note that, these segmentation method generally produce multiple segments. Indeed the graph cut based methods generates hundreds of segments. Hence we prefer methods with higher \emph{F-measure}/\emph{Covering} but using fewer segments. We also report the average number of segments (\emph{Avg num segments}) per image
for each method.
\subsubsection{Performance Comparison}
We compare the object segmentation results of LMSDS with those of other skeleton based methods (Levinshtein's method~\cite{Ref:Levinshtein09}, Lee's method~\cite{Ref:Lee13}, MIL~\cite{Ref:Tsogkas12}
and FSDS~\cite{Ref:ShenCVPR16}), those of graph cut based methods (Shape Sharing~\cite{Ref:Kim12} and CPMC~\cite{Ref:Carreira12}) and that of a deep learning based segmentation
method (FCN~\cite{Ref:LongSD15}). To obtain object segments reconstructed from skeletons, we threshold the thinned skeleton map (after non-maximal suppression) into a binary one.
Thresholds were optimized over the whole dataset according to the F-measures for localization. FSDS does not explicitly predict skeleton scale, but we can estimate a coarse scale
for each skeleton pixel according to the receptive field sizes of the different stages. For each skeleton pixel $x_j$, the scale predicted by FSDS is
$\hat{s}_j=\sum_{i=1}^Mr_i\text{Pr}(z_j=i|X;\mathbf{W}\ast,\mathbf{\Phi}\ast,\mathbf{h}_0\ast)$.
FCN was originally used for semantic segmentation (multi-class classification) in~\cite{Ref:LongSD15}.
Here, we use it for foreground object segmentation (binary classification): Foreground objects have label ``1'' and background have label ``0''.
We finetune the FCN-8s model released in~\cite{Ref:LongSD15} on our datasets to obtain foreground object segmentation.

\begin{figure}[!h]
\centering
\includegraphics[width=1.0\linewidth]{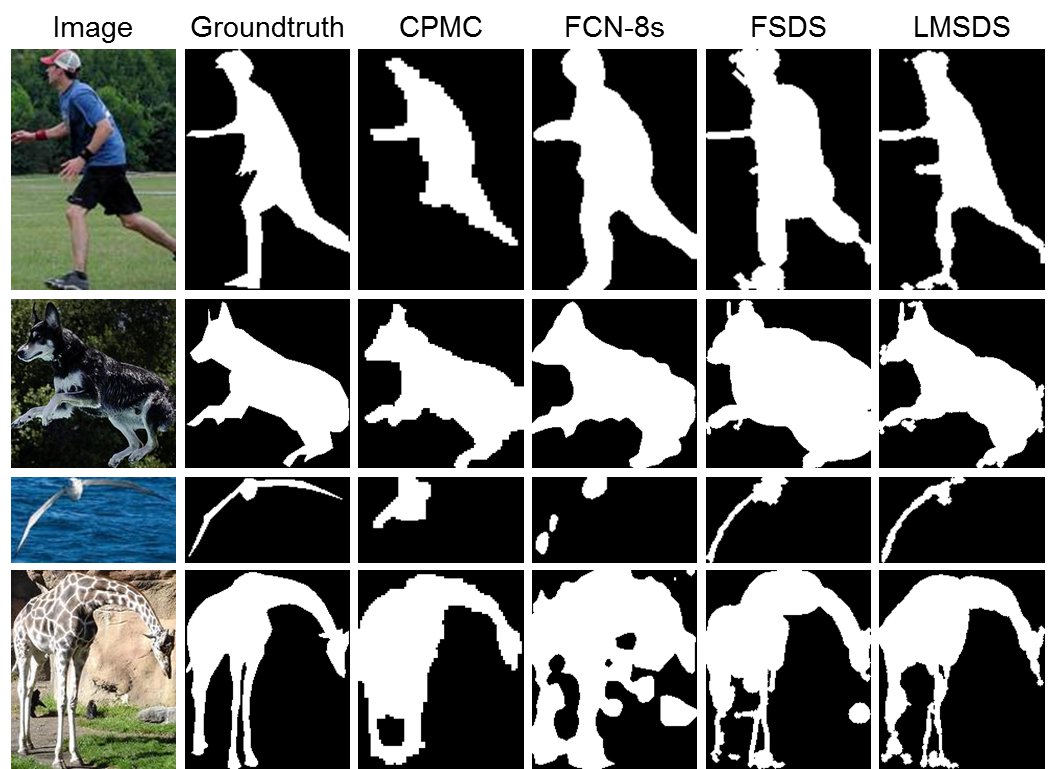}
\caption{Illustration of object segmentation on SK-LARGE for several selected images.
}\label{fig:obj_seg_example_sk}
\end{figure}

\begin{figure}[!h]
\centering
\includegraphics[width=1.0\linewidth]{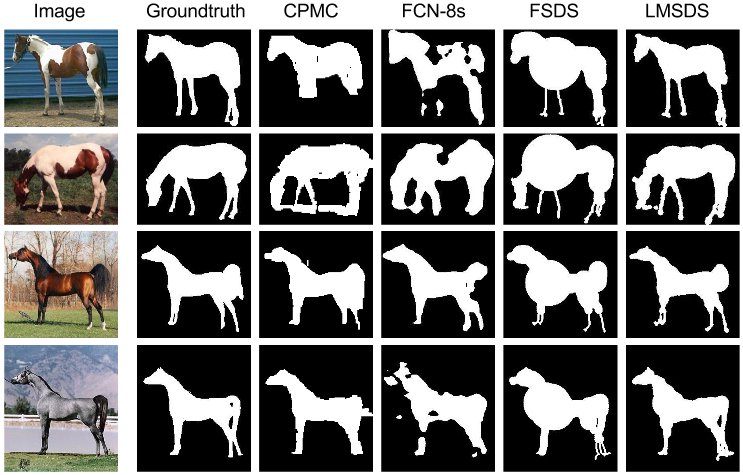}
\caption{Illustration of object segmentation on WH-SYMMAX~\cite{Ref:Shen16} for several selected images.
}\label{fig:obj_seg_example_horse}
\end{figure}

We conduct the object segmentation experiments on SK-LARGE and WH-SYMMAX and evaluate the results according to the segmentation groundtruths provided by
MS COCO~\cite{Ref:Chen15} and Weizmann Horse~\cite{Ref:Borenstein02}, respectively.
The quantitative results on these two datasets are summarized in Table~\ref{tbl:obj_seg_sk_large} and Table~\ref{tbl:obj_seg_horse}, respectively.
LMSDS achieves significant higher \emph{F-measure}/\emph{Covering} than others, except for the result of CPMC on SK-LARGE. However, CPMC has a clear disadvantage compared with LMSDS:
LMSDS only generates about 2 segments per image while CPMC produces 100 times more segments per image, moreover most CPMC segments fires on the background.
Then, as can be seen from the qualitative results illustrated in Fig.~\ref{fig:obj_seg_example_sk} and Fig.~\ref{fig:obj_seg_example_horse}\footnote{Since the graph cut based method (CPMC) generates a large number of segments, we only show the one with the maximum overlap between the groundtruth segment. For others, we show the whole detected foreground segments.},
we find that CPMC misses some significant parts; FCN-8s is usually
unable to ensure smoothness between similar pixels, and spatial and appearance consistency of the segmentation output; FSDS often generates much ``fatter'' bodies due to inaccurate scale predication;
LMSDS produces better segmentation outputs, thanks to the learned scale regressors. Note that even the narrow gap between the tail and the leg of the last horse in Fig.~\ref{fig:obj_seg_example_horse}
can be obtained by LMSDS.

\begin{table}[!h]
\centering
\caption{Object segmentation performance comparison between different methods on SK-LARGE.}\label{tbl:obj_seg_sk_large}
\begin{tabular}{cccc}
\toprule
Method&F-measure&Covering ($\%$)&Avg num segments\\
\midrule
Lee~\cite{Ref:Lee13}&0.496&33.8&210.5\\
MIL~\cite{Ref:Tsogkas12}&0.268&27.5&8.4\\
Shape Sharing~\cite{Ref:Kim12}&0.854&75.4&716.2\\
CPMC~\cite{Ref:Carreira12}&0.896&81.8&287.0\\
FCN-8s~\cite{Ref:LongSD15}&0.840&74.2&3.8\\
FSDS (ours)&0.814&69.1&2.0\\
LMSDS (ours)&0.873&78.1&2.1\\
\bottomrule
\end{tabular}
\end{table}

\begin{table}[!h]
\centering
\caption{Object segmentation performance comparison between different methods on WH-SYMMAX. }\label{tbl:obj_seg_horse}
\begin{tabular}{cccc}
\toprule
Method&F-measure & Covering ($\%$)&Avg num segments\\
\midrule
Lee~\cite{Ref:Lee13}&0.597&43.4&253.0\\
MIL~\cite{Ref:Tsogkas12}&0.278&30.7&8.2\\
Shape Sharing~\cite{Ref:Kim12}&0.857&75.4&879.8\\
CPMC~\cite{Ref:Carreira12}&0.887&80.1&511.2\\
FCN-8s~\cite{Ref:LongSD15}&0.823&72.1&2.3\\
FSDS (ours)&0.838&72.5&1.7\\
LMSDS (ours)&0.902&82.4&1.3\\
\bottomrule
\end{tabular}
\end{table}

\subsection{Object Proposal Detection} \label{sec:detection}
\begin{figure}[!h]
\centering
\includegraphics[width=1.0\linewidth]{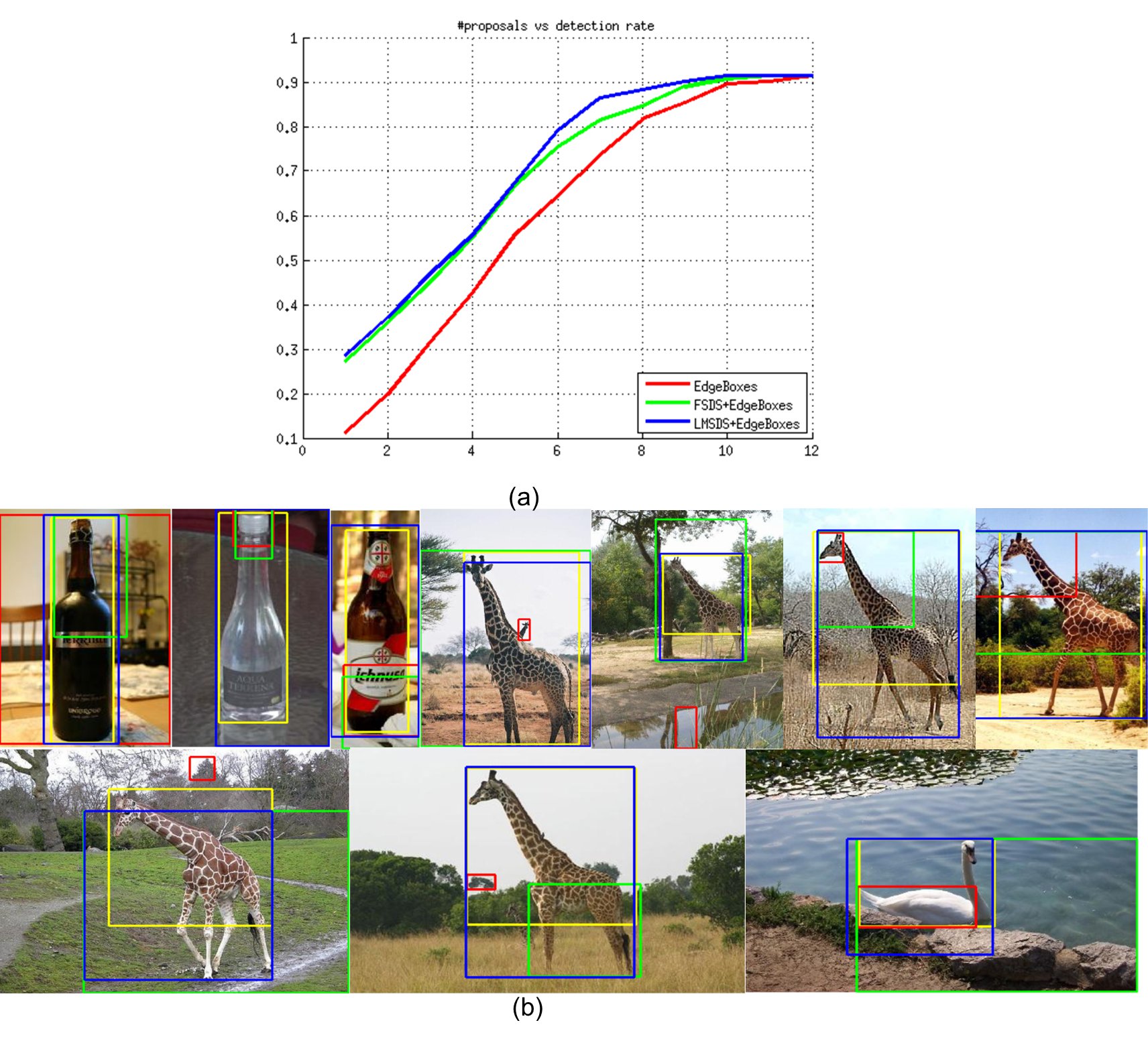}
\caption{Object proposal results on ETHZ Shape Classes~\cite{Ref:Ferrari06}. (a) The detection rate curve (IoU = 0.7). (b) Examples. Groundtruth (yellow), the closest proposal to groundtruth of Edgebox (red), FSDS+EdgeBoxes (green) and LMSDS+EdgeBoxes (blue).
}\label{fig:object_proposals}
\end{figure}

To illustrate the potential of the extracted skeletons for object detection, we performed an experiment on object proposal detection. Let $h_B^E$ be the objectness score of a bounding box $B$ obtained by EdgeBoxes~\cite{Ref:ZitnickD14}, we define our objectness score by $h_B=\frac{\bigcup_{\forall\mathcal {M}{\cap}B\neq\emptyset}(B_{\mathcal {M}}{\cap}B)}{(\bigcup_{\forall\mathcal {M}{\cap}B\neq\emptyset}B_{\mathcal {M}}) \bigcup B}{\cdot}h_B^E$, where $\mathcal {M}$ is a part mask reconstructed by a detected skeleton segment and $B_{\mathcal {M}}$ is the minimal bounding box of $\mathcal {M}$. Let LMSDS+EdgeBoxes and FSDS+EdgeBoxes denote the scoring methods based on the skeletons obtained by LMSDS and FSDS, respectively. As shown in Fig.~\ref{fig:object_proposals}, LMSDS+EdgeBoxes achieves a better object proposal detection result than EdgeBoxes and FSDS+EdgeBoxes.

\section{Conclusion}
We proposed a new network architecture, which is a fully convolutional network with multiple multi-task scale-associated side outputs, to address the unknown scale problem in skeleton extraction. By studying the relationship between the receptive field sizes of the sequential scale-associated side outputs in the network and the skeleton scales they capture, we showed the importance of our proposed scale-associated side outputs for (1) guiding multi-scale feature learning, (2) fusing scale-specific responses from different stages and (3) training with multi-task loss to perform both skeleton localization and scale prediction. The experimental results demonstrate the effectiveness of the proposed method for skeleton extraction from natural images. It achieves significant improvements over the alternatives. We performed additional experiments on applications, such like object segmentation and object proposal detection, which verified the usefulness of the extracted skeletons in object detection.

\section*{Acknowledgment}
This work was supported in part by
the National Natural Science Foundation of China under Grant
61672336 and 61573160, in part by "Chen Guang" project supported by Shanghai Municipal Education Commission and Shanghai                                                        Education Development Foundation under Grant 15CG43, in part by the Intelligence Advanced Research Projects Activity (IARPA) via
Department of Interior/ Interior Business Center (DoI/IBC) contract
number D16PC00007, and in part by Office of Naval Research N00014-15-1-2356. We thank NVIDIA Corporation
for providing their GPU device for our academic research.

\ifCLASSOPTIONcaptionsoff
  \newpage
\fi



\bibliographystyle{IEEEtran}
\bibliography{egbib}
\end{document}